\documentclass[11pt]{article}
\usepackage{nodalida2025}

\usepackage{mathptmx}  

\usepackage{latexsym}

\usepackage{placeins}

\usepackage{amsmath}
\usepackage{amssymb}
\usepackage{multirow}

\usepackage{microtype}

\usepackage{inconsolata}

\usepackage{graphicx}

\usepackage{subcaption}
\usepackage{tikz}
\usetikzlibrary{positioning, shapes.misc}

\usepackage[hidelinks, breaklinks=true]{hyperref}

\newcommand{\src}[1]{{#1}^{src}}
\newcommand{\tgt}[1]{{#1}^{tgt}}

\makeatletter
\def\hlinewd#1{%
\noalign{\ifnum0=`}\fi\hrule \@height #1 \futurelet
\reserved@a\@xhline}
\makeatother

\newcommand{\aligncolor}{teal}

\aclfinalcopy %

\title{Revisiting Projection-based Data Transfer for Cross-Lingual Named Entity Recognition in Low-Resource Languages}

\author{Andrei Politov\thanks{\ \ These authors contributed equally.}\  \thanks{\ \  Corresponding author.},\ Oleh Shkalikov\footnotemark[1],\ René Jäkel,\ Michael Färber \\
\normalsize Center for Scalable Data Analytics and
Artificial Intelligence (ScaDS.AI), \\
TU Dresden, Dresden/Leipzig, Germany\\
\texttt{\{andrei.politov, oleh.shkalikov, rene.jaekel, michael.faerber\}@tu-dresden.de}
}

\date{}

\begin{document}
\maketitle
\begin{abstract}
  Cross-lingual Named Entity Recognition (NER) 
  leverages knowledge transfer between languages 
  to identify and classify named entities, making it particularly useful for low-resource languages.
  We show that the data-based cross-lingual transfer method is an effective technique for cross-lingual NER and can outperform multi-lingual language models for low-resource languages.
  This paper introduces two key enhancements to the annotation projection step in cross-lingual NER for low-resource languages. First, we explore refining word alignments using back-translation to improve accuracy. Second, we 
  present a novel formalized projection approach of matching source entities with extracted target candidates.
Through extensive experiments on two datasets spanning 57 languages, we demonstrated that our approach surpasses existing projection-based methods in low-resource settings. These findings highlight the robustness of projection-based data transfer as an alternative to model-based methods for cross-lingual named entity recognition in low-resource languages.
\end{abstract}

\section{Introduction}

Named Entity Recognition is well-studied in Natural Language Processing (NLP), but remains a challenge for low-resource languages due to the lack of manual annotation \citep{pakhale2023comprehensive}. Of the roughly 7,000 languages spoken worldwide, most are low-resource, with over 2,800 endangered \cite{Ethnologue}. Cross-lingual approaches present a promising solution to address the scarcity of labelled data in these languages. 

Cross-lingual NER methods can be categorized into \textit{model transfer} and \textit{data-based transfer} approaches \citep{garcia-ferrero-etal-2022-model}. \textit{Model transfer approaches} depend on the ability of  multi-lingual models to convey task-specific knowledge across languages. \textit{Data-based methods} automate labelling through translation and annotation projection processes while leveraging advancements in multi-lingual language models to enable zero-shot cross-lingual transfer. This approach allows models trained in high-resource languages to identify and classify named entities in other languages without additional annotated data.
Additionally, categorization can be done through two approaches: translate-test, which labels original sentences in zero-shot settings, and translate-train, which generates labelled data to train a NER model.

Here we contribute to the field of cross-lingual NER by demonstrating the effectiveness of a data-based cross-lingual transfer method that achieves comparable and, in some cases, higher performance of multilingual language models in low- and extremely low-resource language scenarios.

Our work focuses on the projection phase of cross-lingual NER pipelines, introducing two improvements to projection-based methods.
First, we propose a method specifically designed to improve word-to-word alignments. 
Second, we present a novel formalized projection approach of matching source entities with extracted target candidates.
The proposed methods support translate-train and translate-test setups, achieving performance on par with model-based cross-lingual transfer techniques while offering greater flexibility.
We evaluated our approach using the XTREME \citep{ruder-etal-2023-xtreme} and MasakhaNER2 \cite{adelani2022masakhaner20africacentrictransfer} datasets comprising 57 languages in total in translate-test settings.
The source code and the evaluation results are provided in the GitHub repository\footnote{\url{https://github.com/Cross-Lingual-NER/Projection-Data-Transfer-Cross-Lingual-NER}}.
\enlargethispage{\baselineskip}

\section{Related Work}

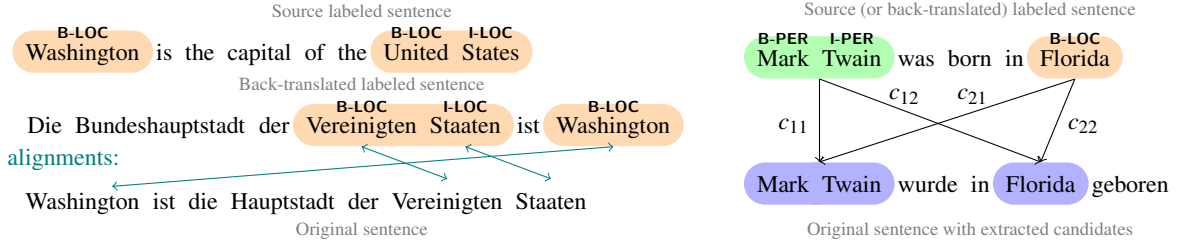
\begin{figure*}[ht]
\begin{subfigure}[t]{0.58\textwidth}
    \centering
    \begin{tikzpicture}[node distance=-0.1,
                font=\footnotesize,
                every node/.style={text centered,
                        text height=1.5ex,
                        text depth=.25ex,},
                loc/.style={fill=orange!30, rounded rectangle,
                        label={[anchor=center,font=\tiny\bfseries\sffamily]above:#1-LOC}}]
            \node[loc={B}](Washington_src){Washington};
            \node[right=of Washington_src](is_src){is};
            \node[right=of is_src](the_src){the};
            \node[right=of the_src](capital_src){capital};
            \node[right=of capital_src](of_src){of};
            \node[right=of of_src](the_src){the};
            \node[loc={B}, rounded rectangle east arc=none, right=of the_src](United_src){United};
            \node[loc={I}, rounded rectangle west arc=none, right=of United_src](States_src){States};

            \node[below=of Washington_src, yshift=-0.5cm, xshift=-0.5cm](Die_backtrans){Die};
            \node[right=of Die_backtrans](Bundeshauptstadt_backtrans){Bundeshauptstadt};
            \node[right=of Bundeshauptstadt_backtrans](der_backtrans){der};
            \node[loc={B}, rounded rectangle east arc=none, right=of der_backtrans](Vereinigten_backtrans){Vereinigten};
            \node[loc={I}, rounded rectangle west arc=none, right=of Vereinigten_backtrans](Staaten_backtrans){Staaten};
            \node[right=of Staaten_backtrans](ist_backtrans){ist};
            \node[loc={B}, right=of ist_backtrans](Washington_backtrans){Washington};

            \node[below=of Washington_src, yshift=-1.5cm](Washington_tgt){Washington};
            \node[right=of Washington_tgt](ist_tgt){ist};
            \node[right=of ist_tgt](die_tgt){die};
            \node[right=of die_tgt](Hauptstadt_tgt){Hauptstadt};
            \node[right=of Hauptstadt_tgt](der_tgt){der};
            \node[right=of der_tgt](Vereinigten_tgt){Vereinigten};
            \node[right=of Vereinigten_tgt](Staaten_tgt){Staaten};

            \node[text=gray, font=\scriptsize, above=of Vereinigten_backtrans, yshift=0.1cm](backtrans){Back-translated labeled sentence};
            \node[text=gray, font=\scriptsize, above=of backtrans, yshift=0.5cm]{Source labeled sentence};
            \node[text=gray, font=\scriptsize, below=of backtrans, yshift=-1.45cm]{Original sentence};
            
            \node[rectangle, 
                text=\aligncolor, 
            ] 
            at (-0.25 cm, -1.4cm)
            {alignments:};

            \draw[<->, \aligncolor] (Washington_tgt.north east)++(-0.5, -0.1) -- (Washington_backtrans.south);
            \draw[<->, \aligncolor] (Vereinigten_tgt.north) -- (Vereinigten_backtrans.south);
            \draw[<->, \aligncolor] (Staaten_tgt.north) -- (Staaten_backtrans.south);
        \end{tikzpicture}
    \caption{Compute word-to-word alignments between back-translated and original sentences}
\end{subfigure}
\hfill
\begin{subfigure}[t]{0.4\textwidth}
        \centering    
        \begin{tikzpicture}[node distance=-0.1,
                font=\footnotesize,
                every node/.style={text centered,
                        text height=1.5ex,
                        text depth=.25ex,},
                loc/.style={fill=orange!30, rounded rectangle, label={[anchor=center,font=\tiny\bfseries\sffamily]above:#1-LOC}},
                per/.style={fill=green!30, rounded rectangle, label={[anchor=center,font=\tiny\bfseries\sffamily]above:#1-PER}},
                cand/.style={fill=blue!30, rounded rectangle},]

            \node[per={B}, rounded rectangle east arc=none](Mark_src){Mark};
            \node[per={I}, rounded rectangle west arc=none, right=of Mark_src](Twain_src){Twain};
            \node[right=of Twain_src](was_src){was};
            \node[right=of was_src](born_src){born};
            \node[right=of born_src](in_src){in};
            \node[loc={B}, right=of in_src](Florida_src){Florida};

            \node[cand, rounded rectangle east arc=none, below=of Washington_src, yshift=-1.2cm](Mark_tgt){Mark};
            \node[cand, rounded rectangle west arc=none, right=of Mark_tgt](Twain_tgt){Twain};
            \node[right=of Twain_tgt](wurde_tgt){wurde};
            \node[right=of wurde_tgt](in_tgt){in};
            \node[cand, right=of in_tgt](Florida_tgt){Florida};
            \node[right=of Florida_tgt](geboren_tgt){geboren};

            \node[text=gray, font=\scriptsize, above=of born_src, yshift=0.2cm](source){Source (or back-translated) labeled sentence};
            \node[text=gray, font=\scriptsize, below=of source, yshift=-2.45cm]{Original sentence with extracted candidates};

            \draw[->] (Mark_src.south east) -- node[left]{\(c_{11}\)} (Mark_tgt.north east);
            \draw[->] (Florida_src.south) -- node[right]{\(c_{22}\)} (Florida_tgt.north);
            \draw[->] (Mark_src.south east) -- node[above left, yshift=0.1cm]{\(c_{12}\)} (Florida_tgt.north);
            \draw[->] (Florida_src.south) -- node[above right, yshift=0.1cm]{\(c_{21}\)} (Mark_tgt.north east);
        \end{tikzpicture}
        \caption{Matching of extracted target candidates with source entities}
    \end{subfigure}
    \caption{Proposed improvements to projection-based cross-lingual NER methods}
    \label{fig:improvments}
\end{figure*}

\textbf{Model transfer} methods leverage the ability of models to transfer task-specific knowledge across languages. For example, multilingual models like mBERT \citep{devlin-etal-2019-bert} and XLM-RoBERTa \citep{conneau-etal-2020-unsupervised} are trained on high-resource languages and applied to low-resource languages without modification. \citet{torge-etal-2023-named} demonstrated improved performance when models were fine-tuned on labelled data or pre-trained on a related language. However, low-resource languages often lack sufficient data, and transfer quality diminishes when applied to very different target languages.

\textbf{Data-based methods} employ labelled datasets, often available in high-resource languages, to perform labelling tasks in the target language. They include fully artificial data generation, like MulDA \citep{liu-etal-2021-mulda}, and annotation projection methods. This paper focuses on the latter, which typically involves three steps: (i) translating the original sentence from the target  (low-resource) to the source (high-resource) language, (ii) applying a NER model to the translated sentence, and (iii)~projecting the labels back to the original sentence. While translation and NER use established models such as BERT \citep{devlin-etal-2019-bert}, many methods have been developed for the projection step.

The first major group \citep{yang-etal-2022-crop, garcia-ferrero-etal-2023-projection, parekh-etal-2024-contextual, Le2024ConstrainedDF} of projection methods is based on back-translation, where labelled source sentences or their parts are translated back to the target language, preserving the labels.
EasyProject \citep{chen-etal-2023-frustratingly} is a translate-train method that employs the insertion of special markers, specifically square brackets, around source entities. 
The marked sentence is then passed to the translation model, which independently translates the entire sentence and each source entity. 
Afterwards, fuzzy string matching is used to project labels: for each substring in the back-translated sentence surrounded by markers, the method identifies the highest fuzzy match for the corresponding translation of the source entity and assigns the appropriate label.

Another type of projection method is based on word-to-word alignments \citep{garcia-ferrero-etal-2022-model, HWA_RESNIK_WEINBERG_CABEZAS_KOLAK_2005, tiedemann-2015-improving, fei-etal-2020-cross, schafer-etal-2022-cross, poncelas-etal-2023-sakura}. 
The general idea is to compute word-to-word correspondence between words of a labelled sentence in a source language 
and an original sentence in a target language.
The entity's label is projected onto target words that align with any of the entity's words.
\citet{garcia-ferrero-etal-2022-model} have shown that using contextualized neural network-based aligners such as SimAlign \citep{jalili-sabet-etal-2020-simalign} or AWESoME \citep{dou-neubig-2021-word} is significantly more beneficial than statistical alignment tools like FastAlign \citep{dyer-etal-2013-simple}, but still can produce wrong alignments and therefore lead to projection errors.

\section{Methodology}

Our proposed approach focuses on projection-based methods that involve 
word-to-word alignments. We present two improvements  (see Figure \ref{fig:improvments}) to existing methods 
which are intended to be useful
for languages that are under-presented in pre-trained language models.
Firstly, we investigate an alternative alignment direction to address the known issue of word-to-word alignment quality.
Secondly, we reformulate the annotation projection task as a bipartite matching problem between source entities and target candidates, using alignment-based matching scores to formalize the problem and eliminate reliance on heuristics, thereby facilitating method extension.

\subsection{Alignment Direction}
In projection-based pipelines, errors can arise at all three stages, diminishing the quality of resulting labels. Handling errors caused by forward translation and source NER models can be challenging. We aim to address projection errors caused by incorrect alignments.

Our approach involves computing word-to-word alignments between the original sentence and its back-translated labelled counterpart in the target language (i.e., target-to-target alignments see Figure \ref{fig:improvments} a). This method is motivated by the expectation that aligning words within the same language is easier than across different languages. This is particularly relevant for low-resource target languages, which often differ significantly from high-resource source languages.

Preserving entities during back-translation is crucial for projecting entities with the use of word alignments between original and back-translated sentences. To achieve this, we employed EasyProject \citep{chen-etal-2023-frustratingly} as outlined in the previous section.

\subsection{Candidate Matching}
The existing methods for addressing problems caused by incorrect alignments such as split annotation, annotation collision and wrong projection fully rely on heuristics \citep{garcia-ferrero-etal-2022-model}.
We consider that the main reason for these issues is a lack of information about any possible entity candidates in the original sentence in a target language. Instead, we propose to generate target entity candidates and match source entities with candidates by solving the weighted bipartite matching problem with additional constraints.

Let \( S \) be a set of source entity spans and \( T \) a set of target candidate spans.
Then \( x_{\src{p}, \tgt{p}} \) is a binary variable which represents whether a source entity \( \src{p} \in S \) is being projected to a target candidate \( \tgt{p} \in T \). Then the source entity-target candidate matching problem can be formulated as follows:
\begin{equation} \label{eq:ent_cand_matching}
    \begin{cases}
        \max\limits_{x} \sum\limits_{\src{p}, \tgt{p} \in S \times T} c_{\src{p}, \tgt{p}} x_{\src{p}, \tgt{p}}           \\
        x_{p_1} + x_{p_2} \leq 1, \quad 
        [\tgt{i}_{p_1},\tgt{j}_{p_1}] \cap [\tgt{i}_{p_2},\tgt{j}_{p_2}] \neq \varnothing \\
        \sum\limits_{\tgt{p} \in T} x_{\src{p}, \tgt{p}} \underset{(\leq)}{=} 1, \quad \forall \src{p} \in S \\
        x_{\src{p}, \tgt{p}} \in \{0, 1\},\quad \forall (\src{p}, \tgt{p}) \in S \times T   
    \end{cases}
\end{equation}
where \( \tgt{p} = ( \tgt{i}_p, \tgt{j}_p ) \in T \) is a candidate span represented as an index of the starting and the ending word, \( c \) is a score of matching. The first set of constraints represents that it is prohibited
to project one or several different source entities to the overlapped candidates. The second ensures that all source entities will be projected.

The generation of target candidates is carried out with either N-grams-
based or NER model-based candidate extraction. The former considers all continuous word sequences as candidates, while the latter predicts the candidate's spans using a multilingual NER model (ignoring predicted classes).

To calculate scores \(c\) from Equation \ref{eq:ent_cand_matching} of matching between source entities and target candidates word-to-word alignments are being used:%
\begin{equation} \label{eq:cost}
    c_{\src{p}, \tgt{p}} = \frac{a_{\src{p}, \tgt{p}}}{\src{j}_p - \src{i}_p + \tgt{j}_p - \tgt{i}_p}
\end{equation}
where \( a_{\src{p}, \tgt{p}} \) is a number of aligned words between a source entity and a target candidate. 
The motivation under this cost is to align entities and candidates based on the count of aligned words, considering source and target lengths to avoid matching with candidates with a lot of nonaligned words and handle single-word misalignments.

The complexity of the proposed problem remains an open question. 
Notably, it is not a straightforward instance of the maximum weight full bipartite matching problem, which can be solved in polynomial time, due to the first set of constraints that prevents projections onto overlapping candidates (i.e. some projections are mutually exclusive). In NER model-based candidate extraction, where no overlapping candidates exist, the problem reduces to a maximum weight bipartite matching. 

To solve the problem in a general formulation, we propose a greedy approximate algorithm, which iteratively selects the projection with the maximum non-zero matching cost, performs this projection, and excludes all candidates that overlap with the projected candidate
as well as the projected source entity.

The proposed concept of target candidate extraction and matching is structurally similar to T-Projection by \citet{garcia-ferrero-etal-2023-projection}, with two key differences. T-Projection uses a fine-tuned T5 model, limiting target languages and producing candidates absent in the original sentence. For matching, T-Projection employs NMTScore by \citet{vamvas-sennrich-2022-nmtscore}, while we use word-to-word alignments.  

\section{Experiments}

We performed an intrinsic evaluation of the efficiency of our approaches across a total of 57 languages using the XTREME \citep{hu2020xtreme} (39 languages) and MasakhaNER2 datasets (excluding Ghom\'al\'a and Naij\'a languages due to limitations in translation model support - 18 languages in total).  This evaluation encompasses the full pipeline, considering both translation and source NER model performance.

\FloatBarrier

\begin{table*}[htpb]
    \centering
   \scalebox{1.15}{
    \footnotesize
    \begin{tabular}{|cc|ccccc|ccc|}
        \hline
        \multirow{2}{*}{\textbf{Approach}}                             & \multirow{2}{*}{\textbf{Align. dir.}} & \multicolumn{5}{c|}{\textbf{XTREME}} & \multicolumn{3}{c|}{\textbf{MasakhaNER2}} \\
        & & yo            & bn  
        & et  & fi & avg & bam & twi & avg   
        \\     
        \hline
        Model transfer              & -      
        & 32.3 & 37.8 & \textbf{67.7} & \textbf{72.1} & \textbf{50.9} & 43.0 & 46.4 & 52.1
        \\
        \hline
        \multirow{2}{*}{Heuristic SimAlign}    & src2tgt          & 32.8          & 36.9          & 56.6 & 59.1 & 41.8 & 49.3 & 71.8 & 66.6  
        \\
                                             & tgt2tgt           & 20.1          & 34.8  
                                             & 39.9 & 50.8 & 34.9 & 44.3 & 5.9 & 42.7
                                             \\
        \hline 
        \multirow{2}{*}{Heuristic AWESoME}    & src2tgt          & 33.3          & \textbf{\underline{38.8}}           & 56.0 & 58.6 & 41.5 & 49.7 & \textbf{\underline{74.6}} & \textbf{\underline{67.3}}   
        \\
                                             & tgt2tgt           & 15.2      & 34.1  
                                             & 40.7 & 50.4 & 34.6 & 43.4 & 3.8 & 42.1
                                             \\
        \hline
        \hline
        \multirow{2}{*}{\textbf{n-gram cand. SimAlign}} & src2tgt           & 29.7 & 36.5 
        & \underline{60.9} & \underline{62.5} & 43.3 & 48.9 & 69.5 & 66.4
        \\
                                             & tgt2tgt  & 17.5          & 36.4          
                                             & 43.4 & 52.5 & 36.4 & 42.5 & 5.2 & 42.4
                                             \\
        \hline
        \multirow{2}{*}{\textbf{n-gram cand. AWESoME}} & src2tgt  & 28.8 & 36.5 
        & 60.0 & 61.7 & 42.3 & 48.3 & 70.9 & 66.7
        \\ 
                                             & tgt2tgt  & 16.1  & 35.2          
                                             & 43.8 & 52.0 & 35.9 & 41.1 & 3.9 & 42.0
                                             \\
        \hline
        \multirow{2}{*}{\textbf{NER cand. SimAlign}} & src2tgt          & \textbf{\underline{52.0}} & 38.3 
        & 58.6 & 61.2 & \underline{46.4} & \textbf{\underline{55.3}} & 69.3 & 63.0
        \\
                                             & tgt2tgt  &  34.0    & 30.7      
                                             & 43.1 & 53.4 & 39.1 & 46.9 & 8.5 & 44.0
                                             \\
        \hline
        \multirow{2}{*}{\textbf{NER cand. AWESoME}} & src2tgt   & 50.2  & 38.2 
        & 58.0 & 60.5 & 45.9 & 55.0 & 69.1 & 62.5
        \\
                                             & tgt2tgt  & 27.7      &         30.7  
                                             & 42.4 & 52.6 & 38.5 & 46.4 & 6.8 & 43.3
                                             \\
        \hline      
    \end{tabular}
   }
    \caption{
    F1 scores for various full pipelines and alignment directions on XTREME (first section) and MasakhaNER2 (second section). \textit{Heuristic SimAlign} and \textit{Heuristic AWESoME} are heuristic approaches, while \textit{n-gram/NER cand. aligner\_name} refers to the proposed candidate matching method with the specified aligner. The first columns show the language where the proposed method outperforms the heuristic the most, the seconds indicate where it underperforms the most, and the last columns provide the average results across all languages. 
\textbf{Bold} values are the overall best, and \underline{underlined} values indicate the best projection-based approaches. Estonian (et) and Finnish (fi) are given as typical examples.
    }
    \label{tab:f1_experiments}
\end{table*}

\makeatletter{\renewcommand*{\@makefnmark}{}
\footnotetext{All models are from the HF Hub}\makeatother}

For a comparative analysis of existing and proposed approaches, we (re)implemented the aforementioned projection methods according to their original papers. 
In particular, we reimplemented the heuristic word-to-word alignment-based approach outlined by \citet{garcia-ferrero-etal-2022-model}. 
We enhanced this heuristic by introducing a word count ratio threshold of 0.8 to better handle misaligned unitary words. 
Additionally, we reimplemented the EasyProject method, which performs back-translation of labelled source sentences, using original, fine-tuned by authors, NLLB-200-3.3B\footnote{\href{https://huggingface.co/ychenNLP/nllb-200-3.3B-easyproject}{ychenNLP/nllb-200-3.3B-easyproject}} model. This back-translated output is then used for annotation projection, relying on word-to-word alignments computed between the original and labelled back-translated sentences in the same language (denoted as \texttt{tgt2tgt}).

NLLB200-3.3B\footnote{\href{https://huggingface.co/facebook/nllb-200-3.3B}{facebook/nllb-200-3.3B}} \citep{nllb200} was employed as a translation model for all experiments.
The XLM-R-Large model\footnote{\href{https://huggingface.co/FacebookAI/xlm-roberta-large-finetuned-conll03-english}{FacebookAI/xlm-roberta-large-finetuned-conll03-english}}, fine-tuned on the English split of the CONLL2003 \citep{tjong-kim-sang-de-meulder-2003-introduction}, served as both the source model and for target candidate extraction, as well as for model transfer experiments. 
We ignored MISC entities predicted by this model in the first set of experiments since this class does not exist in the MasakhaNER2 and XTREME datasets.
For computing word-to-word alignments, we used the original implementations of SimAlign and non-finetuned AWESoME neural aligners with the default settings (with MBERT model).

As the evaluation involved full pipelines, the resulting metrics were influenced by both translation quality and the performance of the NER models. To ensure a fair and consistent comparison of the proposed methods, we employed the same models for translation and source labelling throughout all experiments. For tasks involving the proposed integer linear programming (ILP) formulation of the projection problem, we utilized the previously described greedy approximation algorithm to derive solutions.

Evaluation results for the full pipelines are given in Table \ref{tab:f1_experiments}.

As shown in Table \ref{tab:f1_experiments}, candidate matching methods consistently deliver a strong performance. %
The proposed approach involving n-gram candidates extraction (\textit{n-gram cand.}), compared to heuristics 
(since n-gram does not limit a set of candidates as NER cand. do),
provide comparable or superior results while offering greater flexibility
and avoiding hyperparameter optimization.

The NER model-based extraction (\textit{NER cand.}) generally outperforms model transfer by effectively correcting labels for correctly predicted spans, resulting in greater accuracy particularly when model transfer mislabels these spans. 
It also surpasses the n-gram approach and achieves results comparable to model transfer because of more fine-grained candidates.

The model transfer generally performs better on the XTREME dataset, but candidate matching methods surpass heuristic approaches in most of the 36 languages, except for Bengali, Kazakh, and Swahili. The first may happen due to the model’s exposure to these languages or their partial representations during pretraining, despite being fine-tuned only on English data.

Although the average score for the MasakhaNER2 dataset is modest, the proposed method performs better than heuristics in 10 languages and worse in 8 out of 18 total languages. The full list can be found in the appendix. This discrepancy may be attributed to the simpler morphological structures in the first group(where proposed methods perform better), while the second group, especially languages like Xhosa and Zulu \cite{maho1999comparative}, presents greater morphological complexity, including noun class systems and agreement patterns.

The proposed method with target-to-target alignment direction generally does not outperform the source-to-target method, except for Japanese, due to errors introduced during back-translation, highlighting a potential area for future research.

Additional experiments, described in the appendix, evaluate the performance of the projection step independently. 
Table \ref{tab:f1_europarl} 
shows projection performance on pre-labelled Europarl parallel texts \citep{agerri-etal-2018-building}, excluding translation and source NER labelling errors. It highlights that candidate matching methods yield results comparable to or better than prior approaches.
The NER-based target candidates approach underperforms due to imperfect spans but surpasses plain model transfer by correcting mislabeled spans via source entity projection.

\section{Conclusion}

In this study, we presented novel annotation projection methods based on word-to-word alignments for cross-lingual NER.

The idea to compute word-to-word alignments between the original and back-translated labelled sentences in the same language, aimed at enhancing the quality of these alignments, did not produce the desired outcomes. This approach encountered significant challenges, primarily due to errors that occurred during the back-translation process.

In contrast, the proposed method of extracting candidates and matching them with source entities showed robust results.
More specifically, the proposed formulation generally outperformed previous word-to-word alignment-based projection methods that relied on heuristics to deal with incorrect alignments.

By using the same NER model for candidate extraction as in model transfer, the proposed approach can outperform model transfer. This is achieved by refining the labels for correctly predicted spans through projection from source entities.

Despite its advantages, the proposed approach remains heavily dependent on the quality of word-to-word alignments. However, the formulated ILP problem incorporates these alignments into matching scores that can be combined with other strategies using a weighted sum.

Our findings demonstrate that the projection-based data transfer approach can be a robust alternative to model-based methods for cross-lingual named entity recognition in low-resource languages.

Future research could aim to improve candidate extraction and explore alternative matching costs in addition to the alignment-based one. The proposed formulation, in contrast to heuristic approaches, facilitates the integration of various scoring mechanisms, allowing for the fusion of different scores to effectively address the limitations associated with each individual method.

Moreover, exploring the usage of LLMs for the projection step in cross-lingual NER pipelines shows potential, indicating that the development of multilingual LLMs could help enhance the performance of NER tasks across diverse languages, especially when working with limited labelled data.

\section*{Limitations}

\textit{Translation Model Dependency}: The performance of the proposed methods relies on the quality of the translation model used -- in our case NLLB200-3.3B \citep{nllb200}. Limitations in translation accuracy for certain languages may propagate errors through the pipeline, especially for morphologically complex or resource-scarce languages.

\textit{NER Model Dependency}: 
Models used for extracting candidates or labelling translated sentences in the source language can be a source of errors.
Incorrect predictions or omissions of entities by a model, coupled with the limited capability to correct such errors on the projection step, can adversely affect the quality of the resulting labelling of the original sentence.
In our experiments, we rely on the XLM-R-Large model, fine-tuned on the English split of CONLL2003, although performance metrics may vary with different models.

\textit{Word-to-Word Alignment Model Dependency}: The matching scores in the proposed ILP formulation for the projection step are computed based on word-to-word alignments. 
Therefore, the quality of the projection is inherently bounded by the quality of these alignments.
In our study, we utilized state-of-the-art neural-based alignment models, specifically SimAlign and AWESoME. 
These models surpass previous statistically-based aligners as they incorporate the context of entire sentences. 
However, their performance remains limited. 
Furthermore, the quality of alignments varies between languages, which can be attributed to the representation of languages in the pretraining datasets of the models, as well as the inherent linguistic properties and structural differences among languages.

\textit{Dataset Variability}: The proposed method demonstrates varying effectiveness across datasets, performing well on less complex languages but struggling with those that exhibit higher morphological complexity (e.g., Xhosa and Zulu). This indicates that additional adaptations may be needed for specific linguistic features.

\textit{Generalization Across Languages}: The candidate matching method shows superior performance for most languages but underperforms in specific cases (e.g., Bengali, Kazakh, and Swahili), potentially due to inadequate representation in pre-training.

\textit{Optimization Heuristics}: While the proposed optimization-based projection method reduces reliance on heuristics, the greedy algorithm used to solve the optimization problem may not achieve global optima in all scenarios.

\section*{Acknowledgements}
The authors acknowledge the financial support by the Federal Ministry of Education and Research of Germany and by Sächsische Staatsministerium für Wissenschaft, Kultur und Tourismus in the programme Center of Excellence for AI-research "Center for Scalable Data Analytics and Artificial Intelligence Dresden/Leipzig", project identification number: SCADS24B.

The authors gratefully acknowledge the computing time made available to them on the high-performance computer at the NHR Center of TU Dresden. This center is jointly supported by the Federal Ministry of Education and Research and the state governments participating in the NHR.

Andrei Politov is deeply grateful to his late mother for her unwavering belief in him and the inspiration she provided throughout the development of this research.

~

\bibliographystyle{acl_natbib}
\bibliography{anthology, custom}

\section*{Appendix}
\subsection*{Isolated Evaluation of the Projection Step}

Table \ref{tab:f1_europarl} depicts performance only of the projection step, excluding translation and source NER labeling errors,
on labelled parallel texts from the Europarl-based NER dataset\footnote{\href{https://huggingface.co/datasets/ShkalikovOleh/europarl-ner}{ShkalikovOleh/europarl-ner}} \citep{agerri-etal-2018-building}.
Since our experiments with the tgt2tgt alignment direction yielded negative results, all pipelines presented in the table are only for the src2tgt case.

We can see that annotation projection methods that incorporate candidate matching can achieve results comparable to or better than previous approaches. Specifically, for the German language, the newly proposed method exhibits a significant performance improvement.

\begin{table}[ht]
    \centering
    \begin{tabular}{|c|ccc|}
        \hline
        \textbf{Projection method}                              & \textbf{de}           & \textbf{es}            & \textbf{it}           \\
        \hline
        Heuristic SimAlign   &    80.0     &       \textbf{90.7}  &      87.0     \\
        \hline
        Heuristic AWESoME   &     81.9     &        90.3       &    87.3           \\
        \hline
        \textbf{n-gram SimAlign} &   89.8        & 89.2 & \textbf{87.8}  \\
        \hline
        \textbf{n-gram AWESoME} & \textbf{92.0} & 88.6  & 87.2 \\
        \hlinewd{2.5pt}
        Model transfer &    67.5     & 74.1 & 69.6 \\
        \hline
        \textbf{NER SimAlign} &   74.5  & 79.8 & \textbf{72.3}  \\           
        \hline
        \textbf{NER AWESoME} &    \textbf{74.7}  & \textbf{80.0}  & 72.0 \\
        \hline
    \end{tabular}
    \caption{
    F1 scores resulting from the evaluation of only the projection step using the Europarl-based NER dataset with English as a source language.
    }
    \label{tab:f1_europarl}
\end{table}

The NER-based target candidates approach performs in this experiment worse due to imperfect spans predicted by the model. However, it still outperforms plain model transfer because it corrects wrongly predicted labels for spans using the projection from matched source entities.

In the case of the Spanish language, the heuristic word-to-word alignment-based algorithm slightly outperforms the proposed approach utilizing the n-gram candidate extraction strategy. 
This advantage arises from the algorithm's ability to merge two continuous ranges of target words aligned with source entity words, when only one misaligned word exists between these ranges. 
In contrast, our approach exhibits this capability only in specific situations.

\subsection*{Notes on Complexity of the Problem}
It can be demonstrated that the proposed problem, when excluding the second set of constraints that limit the number of projections for source entities, reduces to the maximum weight independent set problem on interval graphs, which is solvable in polynomial time \citep{PalB96}. 
Therefore, any potential complexity in the entire problem may be due to the combination of non-overlapping constraints and the constraints limiting the number of projections for each source entity. 
Although it is likely that the proposed ILP formulation could be solved in polynomial time, we cannot make a definitive claim since an appropriate algorithm has yet to be identified.

\subsection*{Insights from the MasakhaNER2 Dataset Experiments}

Here, we provide further details on the results across different languages from the MasakhaNER2 dataset.

The set of 10 languages where the proposed method performs better than heuristics from the MasakhaNER2 dataset includes: Bambara (’bam’), Fon (’fon’), Hausa (’hau’), Igbo (’ibo’), Luganda (’lug’), Mossi (’mos’), Shona (’sna’), Swahili (’swa’), Wolof (’wol’), and Yoruba (’yor’). 
The second set of 8 languages where the proposed methods perfom worst includes: Ewe (’ewe’), Kinyarwanda (’kin’), Luo (’luo’), Chichewa (’nya’), Tswana (’tsn’), Twi (’twi’), Xhosa (’xho’), and Zulu (’zul’). The exact metric values can be found in the provided GitHub repo.

\end{document}